\documentclass[twocolumn, switch]{article} 

\usepackage{preprint}

\usepackage{amsmath, amsthm, amssymb, amsfonts}

\usepackage[numbers,square]{natbib}
\usepackage[utf8]{inputenc}	
\usepackage[T1]{fontenc}	
\usepackage{xcolor}		
\usepackage[colorlinks = true,
            linkcolor = purple,
            urlcolor  = blue,
            citecolor = cyan,
            anchorcolor = black]{hyperref}	
\usepackage{booktabs} 		
\usepackage{nicefrac}		
\usepackage{microtype}		
\usepackage{lineno}		
\usepackage{float}			

\usepackage{lipsum}		

\usepackage{newfloat}
\DeclareFloatingEnvironment[name={Supplementary Figure}]{suppfigure}
\usepackage{sidecap}
\sidecaptionvpos{figure}{c}

\usepackage{titlesec}
\titlespacing\section{0pt}{12pt plus 3pt minus 3pt}{1pt plus 1pt minus 1pt}
\titlespacing\subsection{0pt}{10pt plus 3pt minus 3pt}{1pt plus 1pt minus 1pt}
\titlespacing\subsubsection{0pt}{8pt plus 3pt minus 3pt}{1pt plus 1pt minus 1pt}

\usepackage{tikz,xcolor,hyperref}

\definecolor{lime}{HTML}{A6CE39}
\DeclareRobustCommand{\orcidicon}{
	\begin{tikzpicture}
	\draw[lime, fill=lime] (0,0) 
	circle [radius=0.16] 
	node[white] {{\fontfamily{qag}\selectfont \tiny ID}};
	\draw[white, fill=white] (-0.0625,0.095) 
	circle [radius=0.007];
	\end{tikzpicture}
	\hspace{-2mm}
}
\foreach \x in {A, ..., Z}{\expandafter\xdef\csname orcid\x\endcsname{\noexpand\href{https://orcid.org/\csname orcidauthor\x\endcsname}
			{\noexpand\orcidicon}}
}

\title{SA-VLA: Spatially-Aware Flow-Matching for Vision-Language-Action Reinforcement Learning}

\usepackage{xwatermark}
\newwatermark[firstpage,color=gray!90,angle=0,scale=0.28, xpos=0in,ypos=-5in]{*correspondence: \texttt{yu\_xingrui@a-star.edu.sg}}

\usepackage{authblk}

\author[1,2]{Xu Pan\orcidA{}}
\author[3]{Zhenglin Wan}
\author[2\thanks{\tt{yu\_xingrui@a-star.edu.sg}}]{Xingrui Yu\orcidB{}}
\author[1]{Xianwei Zheng\orcidC{}}
\author[1]{Youkai Ke}
\author[4]{Ming Sun}
\author[4]{Rui Wang}
\author[5]{Ziwei Wang\orcidD{}}
\author[2]{Ivor Tsang\orcidE{}}

\affil[1]{State Key Laboratory of Information Engineering in Surveying, Mapping and Remote Sensing (LIESMARS), Wuhan University, Wuhan, P.R. China}
\affil[2]{Centre for Frontier AI Research (CFAR), Institute of High Performance Computing (IHPC), Agency for Science, Technology and Research (A*STAR), Singapore}
\affil[3]{Department of Computer Science, National University of Singapore, Singapore}
\affil[4]{Institute of Information Engineering, Chinese Academy of Sciences, Beijing, P.R. China}
\affil[5]{School of Electrical and Electronic Engineering, Nanyang Technological University, Singapore}

\begin{document}

\twocolumn[ 
  \begin{@twocolumnfalse} 
  
\maketitle

\begin{abstract}
    Vision-Language-Action (VLA) models exhibit strong generalization in robotic manipulation, yet reinforcement learning (RL) fine-tuning often degrades robustness under spatial distribution shifts. 
    For flow-matching VLA policies, this degradation is closely associated with the erosion of spatial inductive bias during RL adaptation, as sparse rewards and spatially agnostic exploration increasingly favor short-horizon visual cues. 
    To address this issue, we propose \textbf{SA-VLA}, a spatially-aware RL adaptation framework that preserves spatial grounding during policy optimization by aligning representation learning, reward design, and exploration with task geometry. 
    SA-VLA fuses implicit spatial representations with visual tokens, provides dense rewards that reflect geometric progress, and employs \textbf{SCAN}, a spatially-conditioned annealed exploration strategy tailored to flow-matching dynamics. 
    Across challenging multi-object and cluttered manipulation benchmarks, SA-VLA enables stable RL fine-tuning and improves zero-shot spatial generalization, yielding more robust and transferable behaviors.
    Code and project page are available at \href{https://xupan.top/Projects/savla/}{https://xupan.top/Projects/savla}
\end{abstract}
\vspace{0.35cm}

  \end{@twocolumnfalse} 
] 



\section{Introduction}
\label{sec:intro}
Enabling robots to perform complex manipulation tasks in the physical world remains a central challenge in embodied artificial intelligence \cite{kemp2007challenges}. 
Robotic manipulation requires not only perceptual recognition but also precise reasoning over object geometry, spatial relations, and contact dynamics to generate reliable actions \cite{gibson2014ecological}. 
Recent Vision-Language-Action (VLA) models demonstrate that large multimodal policies can generalize across diverse tasks by jointly modeling visual observations, language instructions, and action generation \cite{kim2024openvla,kawaharazuka2025vision}. 
In particular, diffusion and flow-matching policies have emerged as a powerful paradigm for continuous, high-dimensional control, enabling multimodal action generation with fine-grained dexterity \cite{DBLP:journals/ijrr/ChiXFCDBTS25}. 
Prominent flow-based VLAs such as $\pi_0$ further highlight the potential of foundation models for robotic manipulation under long-horizon and language-conditioned settings \cite{intelligence2025pi05visionlanguageactionmodelopenworld}.

\begin{figure}[tbp]
\centering
\includegraphics[width=\columnwidth]{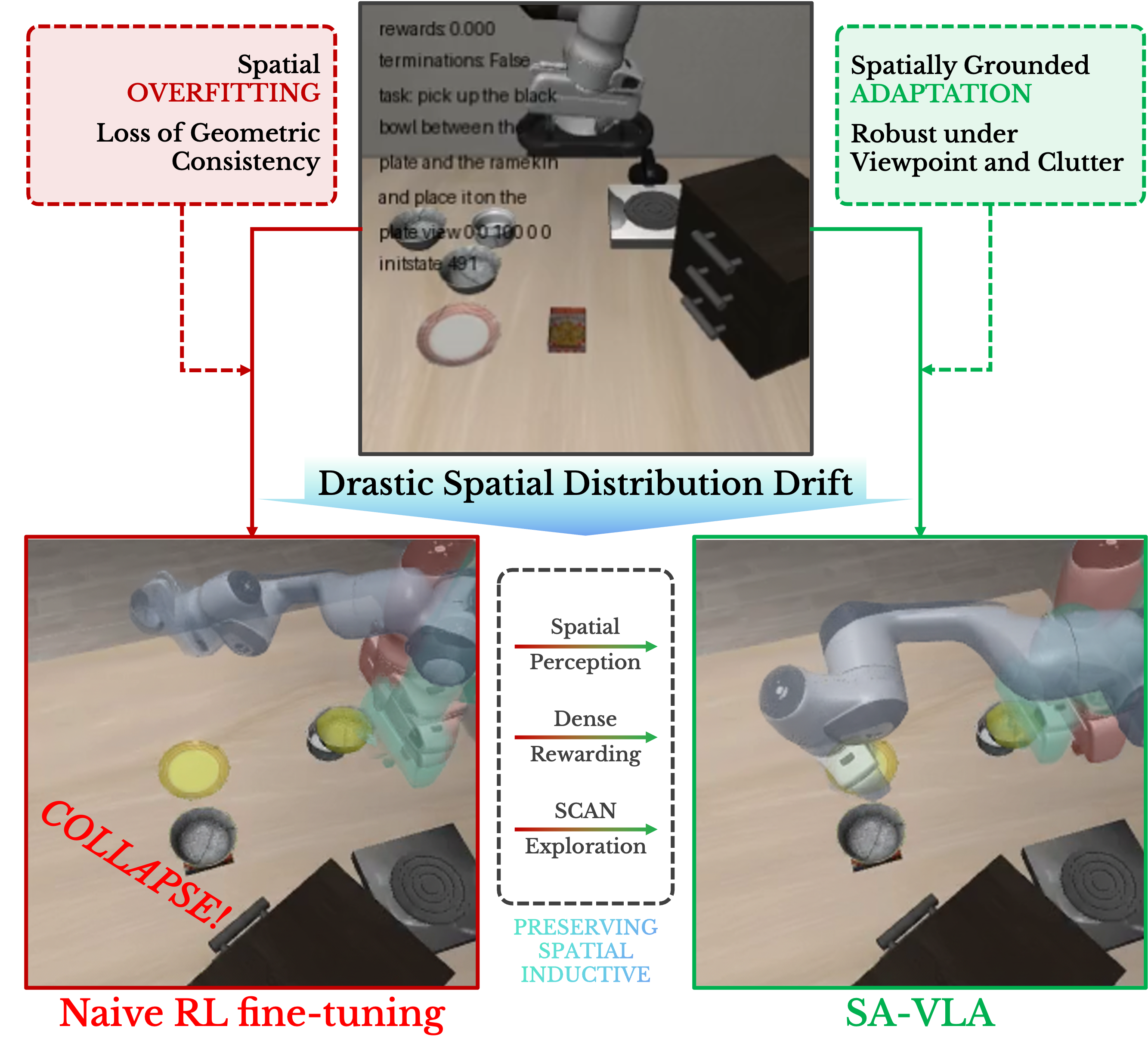}
\caption{
    Illustration of spatial inductive bias collapse during naive RL fine-tuning (left) and preserved spatial grounding with SA-VLA (right) under the same task and identical spatial perturbations.
    For each method, end-effector poses from three temporal phases of a single execution trajectory are rendered as semi-transparent \textcolor[HTML]{FF6F61}{red}, \textcolor[HTML]{50E3C2}{green}, and \textcolor[HTML]{6FA3EF}{blue} masks and overlaid to visualize how spatial behavior evolves over time.
}
\label{fig:intro}
\end{figure}

Despite this progress, adapting pretrained VLA policies via reinforcement learning (RL) remains fragile. 
Although RL fine-tuning \cite{chen2025pi_} can improve in-distribution performance, it often undermines robustness under spatial distribution shifts, such as large viewpoint changes or increased environmental clutter \cite{zhang2025balancing}. 
As shown in Fig.~\ref{fig:intro}, naive RL adaptation can induce phase-inconsistent spatial behavior within a single execution, causing the policy to miss target objects or placement regions under shifted observations. 
These failures arise not solely from the representational limits of visual tokens. 
High-variance updates during on-policy RL can overwrite pretrained geometric regularities.
Meanwhile, sparse or weakly shaped rewards encourage overfitting to spurious reward–action correlations along limited trajectories, reinforcing short-horizon visual cues that fail to generalize across viewpoints or layouts\cite{ma2025comprehensive,tao2025hybrid}.
While spatially agnostic exploration and ambiguous credit assignment are common challenges in RL-based adaptation, their impact is particularly pronounced for flow-matching VLA policies \cite{yin2025deepthinkvla}.
Due to the continuous-time, noise-driven formulation of flow matching, spatial behaviors are maintained largely through implicit geometric priors, which can be gradually eroded under unstable credit assignment, leading to spatial overfitting and reduced geometric consistency under distribution shifts \cite{lyu2025reinforcement}.

To address these challenges, we propose \textbf{SA-VLA}, a spatially-aware framework for RL adaptation of flow-matching Vision-Language-Action policies (Fig.~\ref{fig:intro}). 
SA-VLA stabilizes geometry-aware decision making during RL by injecting spatial structure into state representations, reward signals, and exploration noise.
Here, implicit spatial representations refer to geometric cues derived from 2D visual tokens that support 3D spatial reasoning without requiring 3D reconstruction.
Step-level dense rewards evaluate geometric progress at each interaction, mitigating shortcut learning and improving credit assignment. 
Exploration is guided by \textbf{SCAN} (\textbf{S}patially-\textbf{C}onditioned \textbf{A}nnealed \textbf{N}oise), which modulates action-space noise based on spatial representations while retaining an annealed isotropic diffusion as the underlying reference process.
By combining these elements, SA-VLA reduces catastrophic forgetting and reward-correlation overfitting, ensuring that RL fine-tuning preserves both semantic and spatial consistency under distribution shifts.

The main contributions of this work are as follows:
\begin{itemize}
    \item We introduce \textbf{SA-VLA}, a spatially-aware RL adaptation framework that integrates implicit spatial representations with 2D visual tokens and step-level dense rewards for robust flow-matching VLA policies.
    \item We propose \textbf{SCAN}, a spatially-conditioned exploration strategy that injects geometric understanding into noise-driven policy exploration, improving credit assignment and preserving spatial consistency during RL fine-tuning.
    \item We demonstrate that aligning spatial representations, step-level rewards, and spatially-conditioned exploration yields stable RL fine-tuning and improved generalization under spatial distribution shifts.
\end{itemize}

\section{Related Works}
\label{sec:related}

\textbf{Vision-Language-Action and Spatial Representations.} 
Vision-Language-Action (VLA) models aim to unify visual perception, language instruction, and action generation within a single framework~\cite{ma2024survey, wang2025vlatest}. 
By leveraging large-scale vision-language pretraining and robot demonstrations, these models perform long-horizon, language-conditioned manipulation more generally than task-specific policies~\cite{brohan2022rt, reed2022generalist}. 
Most VLA methods use 2D appearance cues or shallow depth priors and do not explicitly model multi‑view or implicit 3D geometry, limiting geometric continuity and spatial reasoning~\cite{wen2025tinyvla, wang2025vlatest, peng2023openscene}. 
Recent efforts such as BridgeVLA ~\cite{li2025bridgevla} demonstrate strategies for incorporating 3D structure into VLM‑based manipulation learning, highlighting the importance of spatial priors in efficient and generalizable policies. 
StereoVLA~\cite{Deng2025StereoVLAEV} further exploits binocular geometric cues to improve spatial perception and robustness under viewpoint variations. 
A recent survey provides a broader landscape of VLA models and their integration of spatial information.~\cite{shao2025large} 

Explicit 3D representations such as point clouds, voxels, or meshes enable metric reasoning but suffer from discretization artifacts and poor scalability in dynamic or cluttered environments~\cite{qi2017pointnet, dai2017scannet, han2019image, huang2023voxposer}. 
Recent transformer-based backbones such as VGGT~\cite{wang2025vggt} infer high-fidelity implicit spatial tokens directly from multi-view imagery, capturing depth continuity, occlusion, and object topology. 
These implicit spatial representations provide a strong spatial inductive bias beyond 2D visual tokens, enhancing the policy’s ability to reason about spatial relationships, object geometry, and contact dynamics in manipulation tasks.

\begin{figure*}[tbp]
\centering
\includegraphics[width=0.8\linewidth]{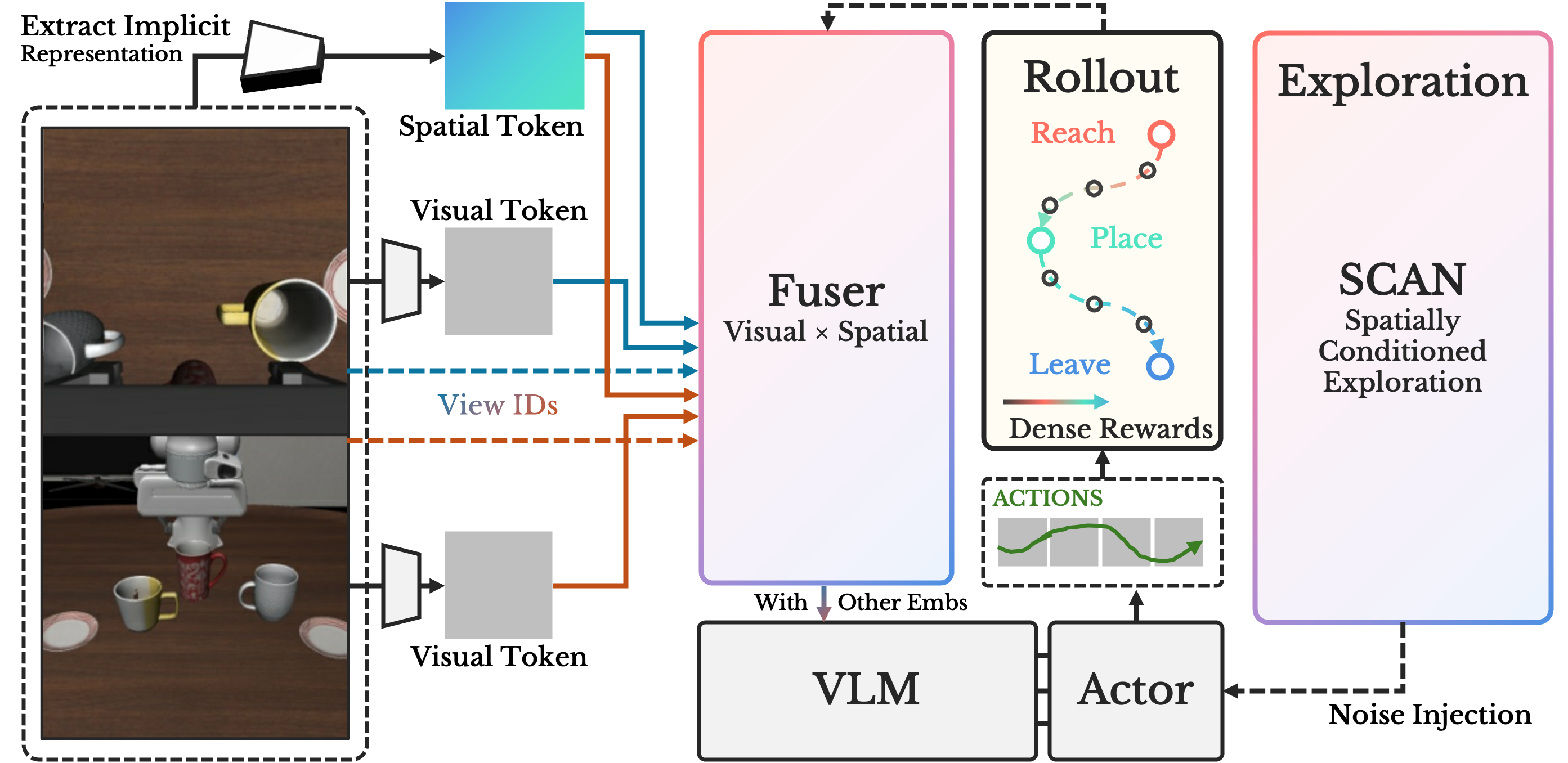}
\caption{
    \textbf{Overview of SA-VLA.}
    Visual and spatial tokens are fused into geometry-aware embeddings, which are optimized via step-level dense rewards and spatially-conditioned exploration (SCAN) for robust RL adaptation.
}
\label{fig:overview}
\end{figure*}

\vspace{3px}
\noindent
\textbf{Multimodal Alignment and Flow-Based Policy Learning.} 
Large vision-language models (VLMs) excel at grounding visual perception to semantic concepts through multimodal pretraining~\cite{alayrac2022flamingo, driess2023palm, zhai2023sigmoid}, yet their embeddings remain largely 2D, lacking consistent spatial structure. 
Efforts incorporating multi-view or depth cues~\cite{peng2023openscene, tang2023emergent} often rely on shallow projections, failing to preserve geometric continuity across viewpoints.

Reinforcement and imitation learning have advanced continuous control and dexterous manipulation~\cite{levine2016end,radosavovic2023real}, but policies often overfit to frequent or easy states due to sparse task-level rewards. 
Intrinsic motivation, self-supervised objectives~\cite{pathak2017curiosity,schwarzer2021pretraining}, and curriculum or difficulty-aware strategies~\cite{portelas2020teacher,racaniere2020automated} partially mitigate this issue, yet most still neglect geometric priors or local spatial difficulty. 
Flow‑matching‑based policy optimization~\cite{zhang2025reinflow, pfrommer2025reinforcement, liu2025flow} improves stability and gradient consistency by matching the gradient flow of optimal trajectories, and recent benchmarks show competitive or superior performance relative to diffusion or RL baselines across manipulation tasks~\cite{gao2025vita}. 
Methods such as ReinFlow further integrate flow matching with online reinforcement learning for stable fine‑tuning of continuous control policies.
Existing methods, however, operate at the trajectory level, are reward-driven, and do not explicitly model step-level progress or leverage implicit spatial representations. 
These limitations motivate our approach, which integrates implicit spatial representations, step-level dense rewards, and spatially-conditioned exploration to robustly learn manipulation policies in cluttered and partially observable environments.

\section{Method}
\label{sec:method}

We propose \textbf{SA-VLA}, a spatially-aware adaptation framework that fuses implicit spatial representations with visual tokens and stabilizes RL via step-level dense supervision and spatially-conditioned exploration (Fig.~\ref{fig:overview}). 
Below, we formalize the problem and describe its three core components: spatial token fusion, step-level dense rewards, and SCAN exploration.

\subsection{Problem Definition}
\label{sec:problem}

We consider adapting pretrained VLA policies to cluttered environments with multiple objects, frequent occlusions, and complex spatial interactions. 
At timestep $t$, the agent observes $s_t = (x_t, o_t)$, where $x_t$ is visual input and $o_t$ includes proprioception or task descriptors. 
Actions $a_t$ are sampled from a policy $\pi_\theta(a_t \mid s_t)$.

Expert demonstrations $\mathcal{D} = \{(x_t, o_t, a_t)\}_{t=1}^N$ may be available, but adaptation relies on RL to maximize expected return:
\begin{equation}
J(\theta) = \mathbb{E}_{\pi_\theta}\Big[\sum_{t=1}^T r(s_t, a_t)\Big],
\end{equation}
where $r(s_t, a_t)$ may provide step-level spatial supervision. 
Naive fine-tuning often degrades pretrained geometric priors, causing brittle behavior under spatial shifts.

\subsection{Method Overview}
\label{sec:overview}

SA-VLA grounds policy updates in geometry. 
Visual tokens $\mathbf{x}_t$ are augmented with implicit spatial tokens $\mathbf{z}_t$, encoding object layout and relative geometry, and fused via
\begin{equation}
\mathbf{h}_t = f_\phi(\mathbf{x}_t, \mathbf{z}_t),
\end{equation}
which conditions the policy $\pi_\theta(a_t \mid \mathbf{h}_t)$.

Policy optimization uses step-level dense rewards $r_t$:
\begin{equation}
\mathcal{L}_\text{RL}(\theta) = - \mathbb{E}_{\pi_\theta}\Big[\sum_t r_t\Big],
\end{equation}
and SCAN exploration injects spatially-aware noise $\epsilon_t^{\text{SCAN}}$, producing actions $a_t = \pi_\theta(\mathbf{h}_t) + \epsilon_t^{\text{SCAN}}$.

The following sections detail spatial token fusion, dense rewards, and SCAN-based exploration.

\subsection{Spatial Token Fusion}
\label{sec:fuser}

\begin{figure}[tbp]
\centering
\includegraphics[width=\columnwidth]{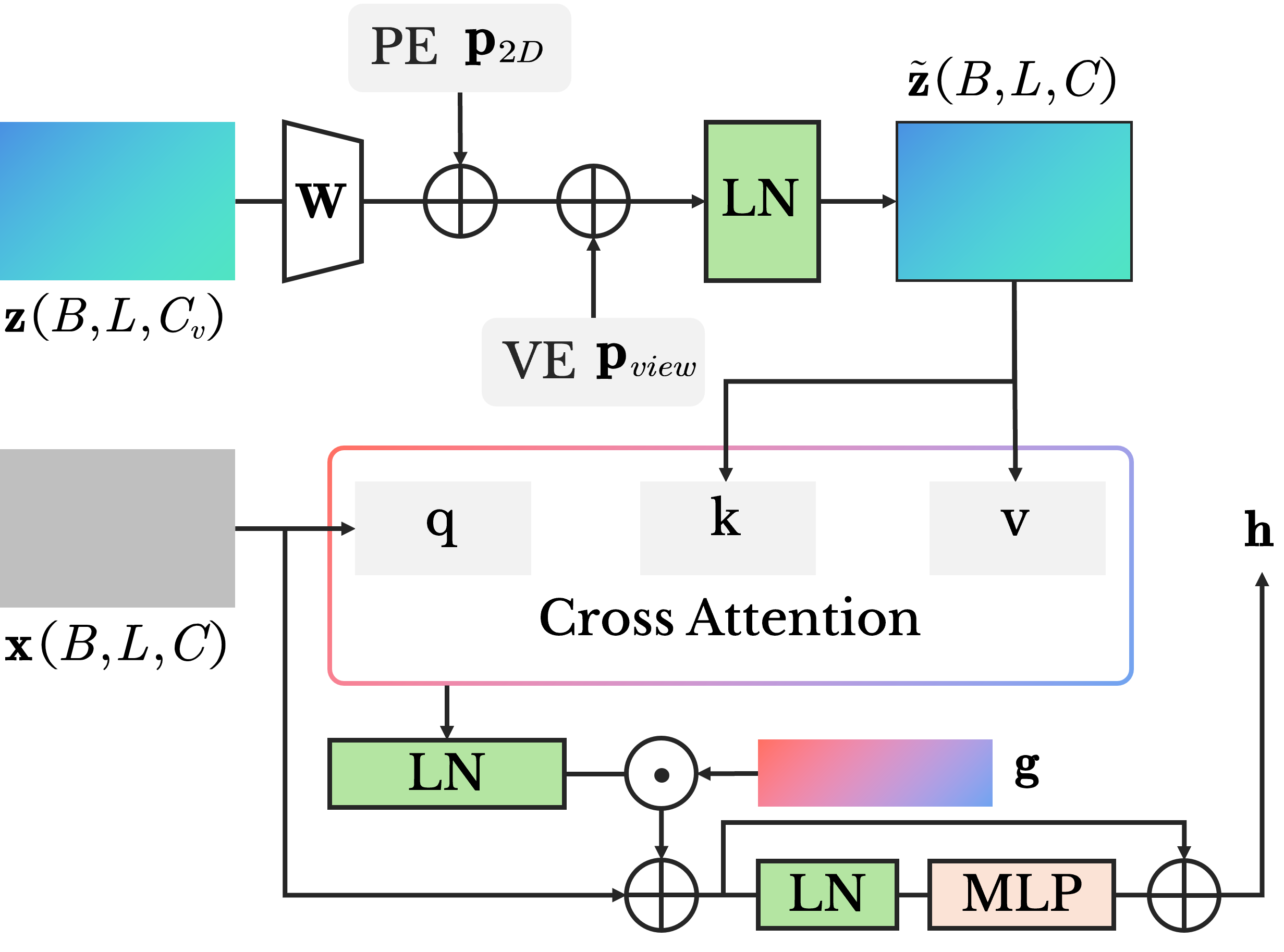}
\caption{
    \textbf{Spatial Token Fusion.}
    Visual semantic tokens $\textbf{x}$ attend to spatial tokens $\textbf{z}$ augmented with positional and view embeddings via unidirectional cross-attention.
    A learnable channel-wise gate $\mathbf{g}$ modulates spatial contributions, followed by a residual MLP to produce fused embeddings for flow-matching policies.
}
\label{fig:fuser}
\end{figure}

Figure~\ref{fig:fuser} illustrates the spatial token fusion module. 
To achieve robust manipulation in spatially complex environments, state representations must capture both fine-grained visual details and underlying 3D structure. 
Visual tokens $\mathbf{x}_t \in \mathbb{R}^{L \times C}$ encode local semantics but are sensitive to occlusion, lighting, and viewpoint changes. 
Complementary geometric cues are provided by implicit spatial tokens $\mathbf{z}_t \in \mathbb{R}^{L \times C_s}$ derived from multi-view features~\cite{wang2025vggt}, which capture scene layout and coarse spatial structure. 
SA-VLA fuses these modalities into unified geometry-aware embeddings for downstream policy conditioning.

Spatial tokens are projected into the visual embedding space and augmented with positional and view encodings:
\begin{equation}
\tilde{\mathbf{z}}_t = \text{LayerNorm}(\mathbf{z}_t \mathbf{W}_\text{proj} + \mathbf{p}_\text{2D} + \mathbf{p}_\text{view}),
\end{equation}
where $\mathbf{p}_\text{2D}$ and $\mathbf{p}_\text{view}$ encode spatial layout and camera biases.

Visual tokens attend to spatial tokens via unidirectional cross-attention:
\begin{equation}
\mathbf{a}_t = \text{CrossAttn}(\mathbf{x}_t, \tilde{\mathbf{z}}_t, \tilde{\mathbf{z}}_t),
\end{equation}
followed by a learnable channel-wise gate $\mathbf{g}$:
\begin{equation}
\mathbf{h}_t = \mathbf{x}_t + \tanh(\mathbf{g}) \odot \text{LayerNorm}(\mathbf{a}_t),
\end{equation}
which stabilizes early RL updates while allowing geometric cues to propagate in occluded regions. 
A residual MLP further refines $\mathbf{h}_t$: $\mathbf{h}_t \gets \mathbf{h}_t + \text{MLP}(\text{LayerNorm}(\mathbf{h}_t))$, preserving pretrained features and producing geometry-aware embeddings.

The resulting $\mathbf{h}_t$ integrates semantic, positional, and multi-view spatial information, forming a robust input for action selection. 
By combining cross-modal attention, adaptive gating, and residual refinement, the fuser mitigates limitations of 2D-only or spatial-only representations. 
These fused embeddings serve as a foundation for the dense reward design described in the next section.

\subsection{Spatially-Aware Step-Level Dense Reward}
\label{sec:dense_reward}

\begin{figure}[!htbp]
\centering
\includegraphics[width=\columnwidth]{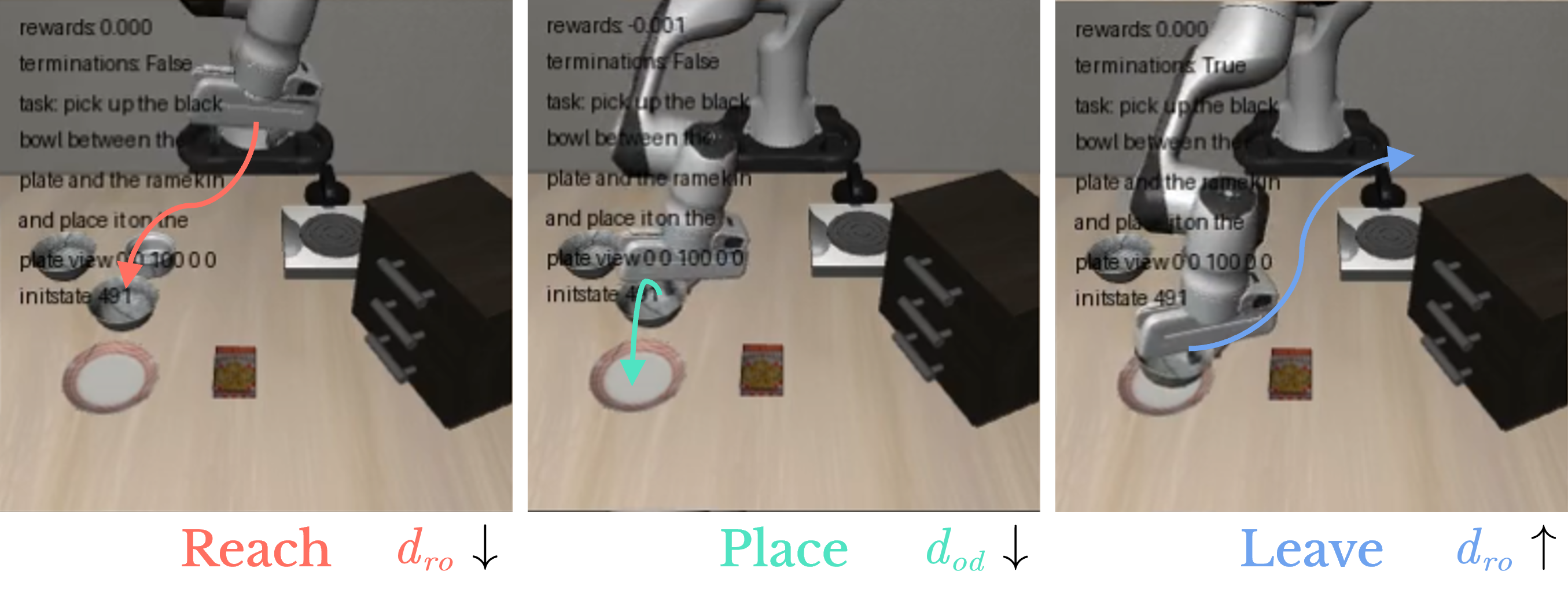}
\caption{
Phase-consistent geometric progress used for step-level dense rewards, decomposing manipulation into Reach, Place, and Leave phases.
}
\label{fig:dense_reward}
\end{figure}

Effective RL in spatially complex manipulation requires feedback that is both dense and geometrically meaningful. 
Sparse, episode-level rewards fail to convey intermediate progress or encode spatial relations critical for robust generalization. 
To address this, SA-VLA introduces a step-level dense reward that provides phase-consistent geometric feedback at each interaction step, explicitly aligned with task structure and manipulation dynamics (Fig.~\ref{fig:dense_reward}).

We decompose a manipulation episode into three semantically distinct phases: \textbf{Reach}, \textbf{Place}, and \textbf{Leave}. 
These phases correspond to generic geometric objectives that naturally arise in object-centric manipulation: approaching the target object, transporting it toward the destination, and disengaging the end-effector after successful placement. 
Unlike manually scheduled trajectory segmentation, this structure enables dense supervision while preserving temporal flexibility during policy execution.

At each step $t$, we compute two normalized geometric distances. 
Let $\mathbf{p}_\text{eef}$, $\mathbf{p}_\text{obj}$, and $\mathbf{p}_\text{dest}$ denote the end-effector, target object, and destination positions, respectively. 
We define:
\begin{equation}
d_{ro}^{(t)} = \frac{\lVert \mathbf{p}_\text{eef}^{(t)} - \mathbf{p}_\text{obj}^{(t)} \rVert}{d_{ro}^{\mathrm{ref}}},\quad
d_{od}^{(t)} = \frac{\lVert \mathbf{p}_\text{obj}^{(t)} - \mathbf{p}_\text{dest}^{(t)} \rVert}{d_{od}^{\mathrm{ref}}},
\end{equation}
where $d_{ro}^{\mathrm{ref}}$ and $d_{od}^{\mathrm{ref}}$ are reference distances measured at the first valid interaction step and used to normalize subsequent measurements. 
Distances are clipped to $[0, 1]$ to ensure numerical stability.

The dense reward is formulated as a \emph{phase-specific temporal progress signal}, computed from the signed change of the relevant normalized distance:
\begin{align}
r_t^\text{Reach} &= \lambda \big(d_{ro}^{(t-1)} - d_{ro}^{(t)}\big), \\
r_t^\text{Place} &= \lambda \big(d_{od}^{(t-1)} - d_{od}^{(t)}\big), \\
r_t^\text{Leave} &= \lambda \big(d_{ro}^{(t)} - d_{ro}^{(t-1)}\big),
\end{align}
where $\lambda$ is a scaling coefficient. 
Reach and Place reward reductions in end-effector-object and object-destination distances, while Leave rewards increasing separation between the end-effector and object to discourage post-placement interference.

Phase assignment is inferred online from low-level interaction signals rather than predefined temporal boundaries. 
Transitions are triggered based on gripper state stability and relative object poses across consecutive steps, combined with task-relevant thresholds such as object-destination proximity. 
A priority-based rule enforces mutually exclusive phase selection: Leave overrides Place once the object is released and stably positioned, while rapid oscillations between phases are prevented. 
This stability-driven inference ensures consistent phase attribution without relying on annotations or fixed temporal segments.

By converting geometric task structure into immediate, phase-consistent learning signals, step-level dense rewards embed manipulation priors directly into the RL objective. 
While the Reach-Place-Leave decomposition reflects common manipulation semantics, the reward formulation itself depends only on relative geometric progress and online phase inference, making it adaptable to different objects and task variants. 
Combined with fused spatial-visual embeddings, this reward design enables stable fine-tuning of flow-matching VLA policies, preserves zero-shot generalization, and enforces both local and global spatial consistency in complex, cluttered environments.

\subsection{Spatially-Conditioned Annealed Noise (SCAN)}
\label{sec:scan}

The preceding sections establish two key ingredients for robust RL adaptation in spatially complex manipulation: 
geometry-aware state representations via spatial token fusion and informative learning signals via step-level dense rewards.
However, effective exploitation of these signals critically depends on exploration.
Without sufficiently diverse and geometry-consistent trajectories, even well-shaped rewards provide limited supervision, particularly under occlusion, contact uncertainty, and multi-modal action distributions.
This motivates a spatially-aware exploration mechanism that is aligned with the same geometric structure encoded in the policy input and reward design.

To this end, we introduce \textbf{SCAN} (\emph{Spatially-Conditioned Annealed Noise}), an exploration strategy that injects stochasticity conditioned on spatial embeddings while enforcing an annealed minimum noise floor.
SCAN ensures persistent exploration throughout fine-tuning, avoids premature collapse to deterministic behaviors, and respects the local geometry of the manipulation space.

We treat the flow-matching policy as defining a stochastic action distribution that is fully compatible with PPO’s surrogate objective. We consider a standard Markov decision process $(\mathcal{S}, \mathcal{A}, P, r, \gamma)$, where $s_t \in \mathcal{S}$ denotes the state, $a_t \in \mathcal{A}$ the action, $P$ the transition kernel, $r$ the reward function, and $\gamma$ the discount factor.
Policy optimization aims to maximize the expected return
\begin{equation}
J(\theta) = \mathbb{E}_{\pi_\theta} \Big[ \sum_{t=0}^{T} \gamma^t r(s_t, a_t) \Big].
\end{equation}

Flow-matching VLA policies model continuous, high-dimensional control dynamics as
\begin{equation}
x_{t+\delta} = x_t + \delta\, f_\theta(x_t,t) + g_t(x_t)\, \epsilon, 
\quad \epsilon \sim \mathcal{N}(0, I),
\end{equation}
where $f_\theta$ predicts the deterministic flow and $g_t(x_t)$ scales the stochastic component.
In practice, unconstrained learning of $g_t(x_t)$ often leads to noise collapse in low-gradient or locally stable regions, yielding nearly deterministic trajectories that hinder exploration.
Conversely, fixed isotropic noise ignores spatial structure and may induce inefficient or unsafe behaviors.
SCAN addresses this trade-off by coupling learned, geometry-aware noise with an annealed lower bound.

\begin{algorithm}[tb]
  \caption{SCAN: Spatially-Conditioned Annealed Noise}
  \label{alg:scan}
  \begin{algorithmic}
    \STATE {\bfseries Input:} policy $\pi_\theta$, dataset of states $s_t$, annealing schedule $\sigma_\text{min}(t)$
    \FOR{each training step $t$}
      \STATE Extract spatial features $x_t$ from $s_t$
      \STATE Predict learned noise $\sigma_\text{learned}(x_t)$
      \STATE Compute total noise: \\$\sigma_t(x_t) = \sigma_\text{min}(t) + h(\sigma_\text{learned}(x_t) - \sigma_\text{min}(t))$
      \STATE Sample $\epsilon_t \sim \mathcal{N}(0, \sigma_t^2(x_t))$
      \STATE Inject noise: $x_{t+\delta} = x_t + \delta f_\theta(x_t,t) + g_t(x_t) \, \epsilon_t$
      \STATE Update policy via PPO using perturbed trajectories
    \ENDFOR
  \end{algorithmic}
\end{algorithm}

Specifically, SCAN defines the effective noise scale as
\begin{equation}
\sigma_t(x_t) = \sigma_\text{min}(t) + h\big(\sigma_\text{learned}(x_t) - \sigma_\text{min}(t)\big),
\end{equation}
where $\sigma_\text{learned}(x_t)$ is predicted from the fused visual-spatial embedding $\mathbf{h}_t$,
$h(\cdot)=\log(1+\exp(\cdot))$ is a smooth monotonic function,
and $\sigma_\text{min}(t)$ is an annealed minimum floor that prevents vanishing stochasticity.
The floor is scheduled as
\begin{align}
\sigma_\text{min}(t) &= \alpha(t)\sqrt{\frac{t}{1-t}}, \\
\alpha(t) &= \alpha_0 + (\alpha_1 - \alpha_0)\frac{\min(k,K)}{K},
\end{align}
where $k$ denotes the global training step and $K$ the annealing horizon.
This construction enforces strictly positive noise throughout training, effectively maintaining a lower bound on policy entropy and avoiding premature convergence.

Formally, SCAN induces a spatially-adaptive stochastic process
\begin{align}
dx &= f_\theta(x,t)\,dt + \Sigma(x,t)\,dW_t, \\
\Sigma(x,t) &= \mathrm{diag}(\sigma_t^2(x)), \\
\Sigma(x,t) &\succeq \Sigma_\text{min}(t),
\end{align}
where $dW_t$ denotes a Wiener process~\cite{song2020score}.
The learned component $\sigma_\text{learned}(x_t)$ captures local geometric and perceptual uncertainty, assigning higher variance in regions of contact sensitivity, spatial ambiguity, or incomplete observation, while allowing near-deterministic behavior in well-explored states.
The annealed floor $\sigma_\text{min}(t)$ encourages broad exploration early and gradually shifts toward fine-grained refinement, implementing a coarse-to-fine homotopy over the spatial action landscape.

By conditioning exploration noise on the same geometry-aware embeddings used for policy prediction and by enforcing an annealed minimum, SCAN aligns exploration with step-level dense rewards.
This synergy increases the likelihood of visiting geometrically meaningful states where phase-wise rewards provide informative gradients, improving credit assignment across manipulation phases.
When integrated with PPO, SCAN balances exploration and exploitation in complex, cluttered environments, stabilizes online fine-tuning, and reinforces behaviors that generalize under viewpoint shifts and occlusion.

Algorithm~\ref{alg:scan} summarizes the SCAN procedure, highlighting spatial feature extraction, noise computation, and policy updates with spatially-conditioned stochasticity.

\section{Experiments}
\label{sec:experiments}

This section evaluates SA-VLA under spatial distribution shifts, with a focus on robustness and training stability during reinforcement learning adaptation.
Rather than only measuring performance gains, we aim to attribute robustness improvements to specific design choices in representation, reward design, and exploration.

Accordingly, we structure the experimental analysis around three research questions:

\begin{itemize}
    \item \textbf{RQ1:} Does explicit spatial token fusion improve zero-shot spatial generalization before RL adaptation?
    \item \textbf{RQ2:} How does reward density influence robustness and optimization stability under noise-injected RL?
    \item \textbf{RQ3:} Does spatially-conditioned exploration provide additional benefits beyond dense rewards when spatial coverage is limited?
\end{itemize}

Each question isolates a distinct component of SA-VLA, enabling a component-wise analysis of spatial inductive bias across representation learning, reward shaping, and exploration.

\subsection{Datasets}

We conduct experiments on LIBERO~\cite{liu2023libero} and its robustness extension LIBERO-PLUS~\cite{fei25libero-plus}. 
LIBERO consists of language-conditioned manipulation tasks with visual and proprioceptive observations, while LIBERO-PLUS introduces systematic perturbations in camera pose and initial state. 
These perturbations evaluate whether a policy preserves consistent spatial reasoning under changes in viewpoint and configuration, a regime where RL fine-tuning has been observed to degrade spatial inductive bias (Sec.~\ref{sec:intro}).

For few-shot RL, we construct a sparse spatial subset by uniformly sampling three trajectories per task and perturbation pair, resulting in 60 pairs. 
Evaluation is performed using 80 parallel environments over five non-repeating sampling epochs to reduce variance and ensure stable comparison across methods.

\subsection{Implementation Details}

All experiments initialize from a pretrained flow matching VLA policy with a frozen visual language backbone. 
Implicit spatial tokens are extracted using a pretrained multi view spatial encoder and fused with visual tokens as described in Sec.~\ref{sec:fuser}. 
The resulting embeddings $\mathbf{h}_t$ condition the policy during action generation.

Policy optimization uses actor-critic PPO with GAE. 
The learning rates are $5\times10^{-6}$ for the policy and $1\times10^{-4}$ for the value function, with a clip ratio of 0.2, discount factor $\gamma=0.99$, and GAE $\lambda=0.95$. 
We apply gradient clipping at 1.0 and use a global batch size of 1024 aggregated from 64 parallel environments over eight rollout epochs. 
The maximum episode length is 240 steps, and training runs for 100 total steps, requiring approximately 38.5 hours on four H800 GPUs.

Sparse reward agents receive only terminal success signals. 
Dense reward agents additionally use step-level spatial rewards defined in Sec.~\ref{sec:dense_reward}, with the dense reward coefficient set to $\lambda_\text{dense}=0.3$. 
Exploration noise is injected using SDE~\cite{liu2025flow}, learnable flow noise~\cite{zhang2025reinflow}, or the proposed SCAN described in Sec.~\ref{sec:scan}, with an annealing threshold of 80 steps.

Evaluation is conducted using 80 parallel environments over five non repeating sampling epochs, with two random seeds per setting. 
We found trends to be consistent across seeds, and report standard deviations to reflect remaining variance.
All metrics report mean and standard deviation.

\subsection{Zero-Shot Spatial Generalization Enabled by Spatial Token Fusion}
\label{sec:exp_spatial_module}

We first address \textbf{RQ1} by isolating the effect of spatial representations before RL adaptation.
All models are trained on the full set of LIBERO tasks using supervised learning and evaluated zero-shot on the spatial-perturbation subset of LIBERO-PLUS. 
This evaluation subset includes variations in camera viewpoint and object initial configurations, testing the ability of policies to generalize without additional training.

Table~\ref{tab:spatial_module_zero_shot} shows that incorporating spatial token fusion improves success rates across all perturbation types. 
The overall gain of $+2.25\%$ demonstrates that explicitly encoding geometric structure strengthens task-relevant spatial reasoning under distribution shifts. 
The improvement is more pronounced under camera-view perturbations ($+3.83\%$) than under initial-state perturbations ($+0.52\%$), indicating that multi-view spatial cues enhance robustness to observation-level changes while offering limited benefit for initial-state variations.

\begin{table}[t]
    \centering
    \caption{
    Zero-shot success rate (\%) on the spatial-perturbation subset of LIBERO-PLUS.
    All models are trained on LIBERO and evaluated without RL adaptation.
    }
    \label{tab:spatial_module_zero_shot}
    \begin{tabular}{lccc}
    \toprule
    Method & View & Init State & Total \\
    \midrule
    $\pi_{0.5}$ (baseline) & 78.47 & 83.77 & 81.00 \\
    Ours (w/ Spatial)      & \textbf{82.30} & \textbf{84.29} & \textbf{83.25 (+2.25)} \\
    \bottomrule
    \end{tabular}
\end{table}

These results establish a controlled baseline before RL adaptation, isolating the contribution of spatial token fusion. 
They provide a reference point for evaluating how reward design and exploration strategies further improve robustness under spatial distribution shifts.

\subsection{Effect of Reward Density on Spatial Generalization}
\label{sec:exp_reward_density_spatial}

\begin{figure}[t]
    \centering
    \includegraphics[width=0.95\columnwidth]{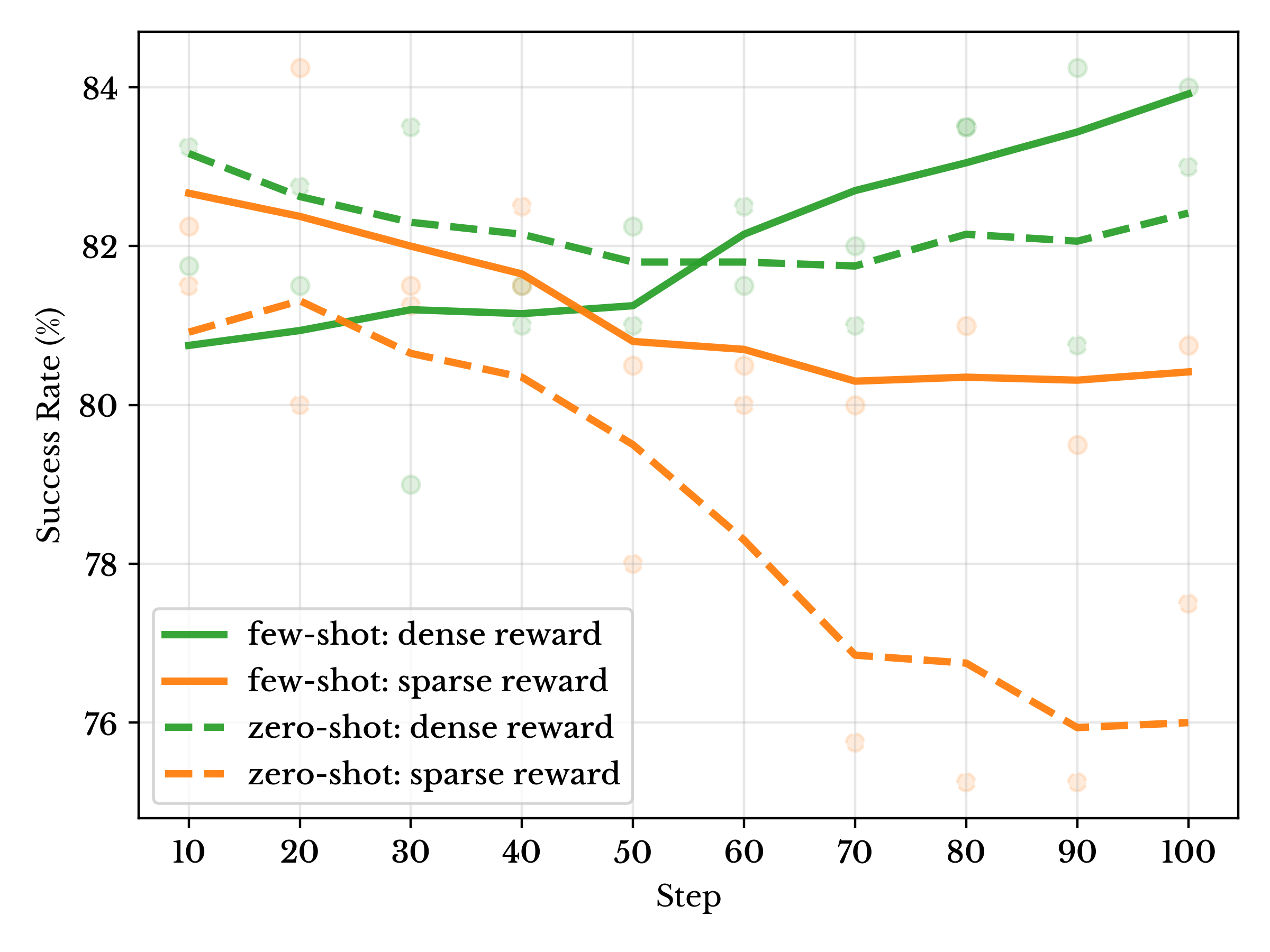}
    \caption{
        \textbf{Training dynamics on the LIBERO-PLUS spatial-perturbation subset.}
        Success rates are evaluated using SDE-based policy checkpoints saved every 10 training steps.
        Solid curves denote few-shot RL, and dashed curves denote zero-shot evaluation.
        Zero-shot evaluation uses 8 environments with a global batch size of 384, while few-shot RL uses 64 environments with a global batch size of 2048.
    }
    \label{fig:reward_density_spatial}
\end{figure}

We next examine \textbf{RQ2}, studying how reward density influences spatial generalization during noise-injected RL fine-tuning. 
This evaluation addresses whether optimization stability under spatial perturbations depends on supervision granularity. 
All experiments use a fixed spatial-perturbation subset of LIBERO-PLUS and apply the same SDE-based stochasticity for policy exploration.

Figure~\ref{fig:reward_density_spatial} compares sparse and dense reward signals under zero-shot and few-shot evaluation. 
Dense rewards consistently lead to faster convergence and smoother training dynamics across all settings. 
Notably, this effect is observed even in zero-shot evaluation, where no task-specific adaptation occurs, indicating that dense rewards stabilize learning rather than solely improving final performance.

The difference arises from how rewards constrain noise-induced exploration. 
Sparse rewards provide weak guidance for credit assignment, leading to high-variance gradients and unstable updates. 
Dense rewards supply step-level supervision aligned with intermediate spatial progress, guiding exploration toward meaningful states and regularizing policy updates under stochastic perturbations.

This stabilizing effect is especially pronounced in few-shot settings, where limited trajectory coverage amplifies optimization noise. 
Dense supervision prevents overfitting to incidental behaviors from stochastic exploration and preserves consistent spatial reasoning across perturbations.

Overall, these results show that reward density is a key factor for robust RL adaptation under spatial distribution shifts, complementing spatial representations by improving optimization stability rather than only enhancing final success rates.

\subsection{Spatially-Conditioned Exploration Beyond Dense Rewards}
\label{sec:exp_scan_rl}

\begin{figure}[t]
    \centering
    \includegraphics[width=0.95\columnwidth]{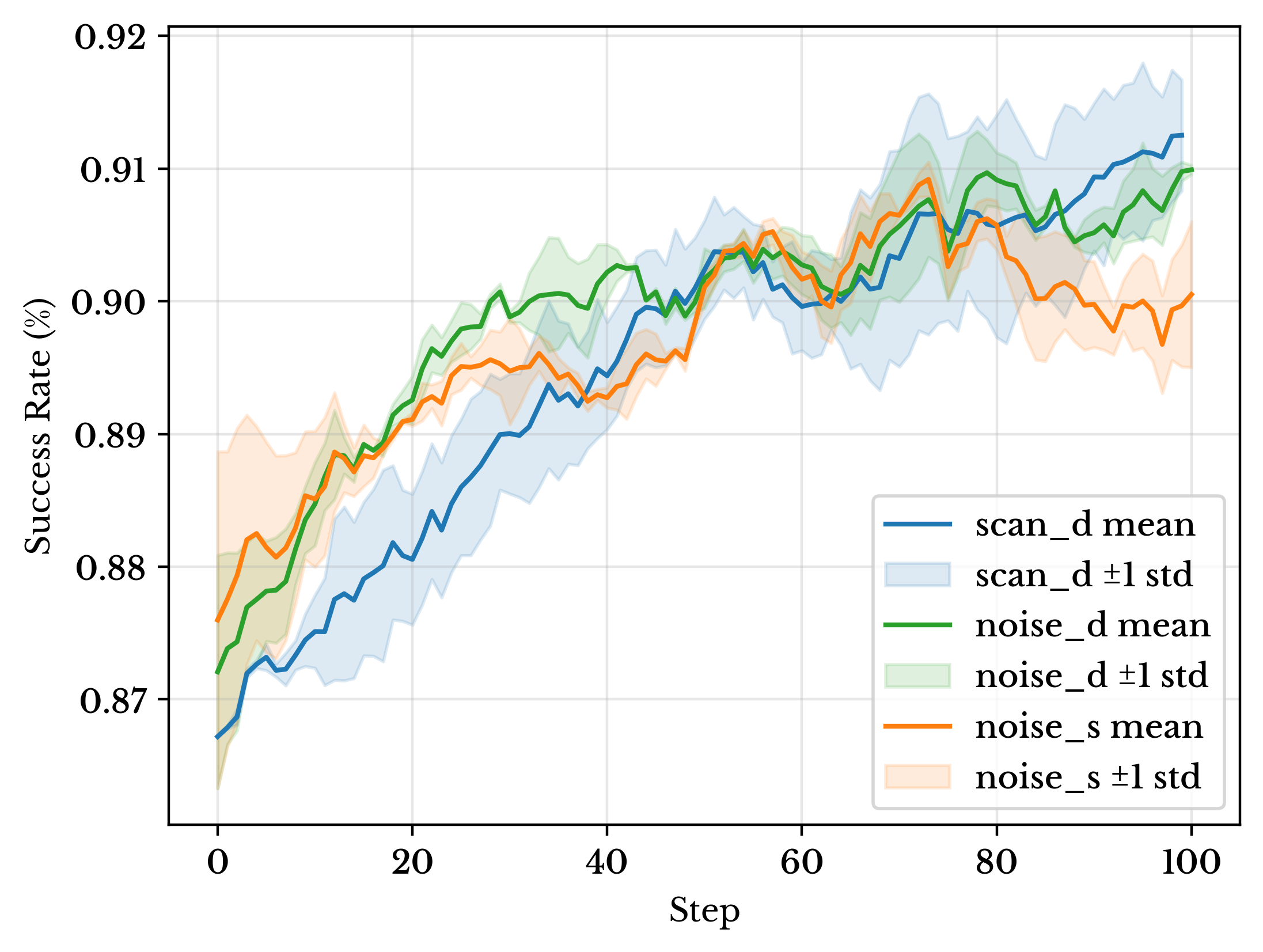}
    \caption{
        \textbf{Training dynamics under limited spatial coverage.}
        Dense rewards stabilize RL optimization, while combining dense rewards with SCAN further improves final success rate.
        Shaded regions denote one standard deviation over two seeds.
    }
    \label{fig:scan_training_curves}
\end{figure}

While step-level dense rewards stabilize RL optimization (Sec.~\ref{sec:exp_reward_density_spatial}), limited spatial coverage can still hinder the policy from encountering key geometric configurations. 
This motivates \textbf{RQ3}, assessing whether spatially-conditioned exploration via SCAN further preserves spatial inductive bias during fine-tuning.

We evaluate three regimes: 
(1) flow\_noise~\cite{zhang2025reinflow} with sparse rewards, 
(2) flow\_noise with dense rewards, and 
(3) dense rewards combined with SCAN.
Figure~\ref{fig:scan_training_curves} illustrates training dynamics under few-shot RL on spatially perturbed environments.

Sparse rewards alone lead to unstable updates and local convergence, as the policy receives weak guidance on intermediate geometric progress. 
Dense rewards reduce gradient variance and stabilize optimization, but in settings with limited trajectory coverage, the agent may fail to visit critical states, resulting in partial spatial exploration and suboptimal generalization. 
In contrast, SCAN injects spatially-conditioned stochasticity aligned with the fused visual-spatial embeddings, actively guiding the agent to under-sampled geometric regions while maintaining consistency with phase-specific dense rewards.

The combination of dense rewards and SCAN yields three key effects: 
(1) gradients are well-aligned with geometric progress, preventing reward-action shortcuts; 
(2) exploration covers spatially relevant states that are otherwise missed, improving state visitation and credit assignment; 
(3) stochasticity preserves controlled diversity, reflected in higher variance that corresponds to targeted, spatially-aware exploration rather than training instability.

As a result, SCAN enables policies to maintain robust spatial reasoning under distribution shifts, achieving higher final success rates than dense rewards alone. 
These observations support our central claim that coupling geometry-aware dense rewards with spatially conditioned exploration preserves the spatial inductive bias of flow-matching VLA policies during RL adaptation, leading to stable and transferable behaviors under spatial perturbations.

\subsection{Ablation Study}
\label{sec:ablations}

We perform a cumulative ablation study to quantify the contribution of each spatial component in SA-VLA. 
All experiments are evaluated on the LIBERO-PLUS spatial-perturbation subset, and unless noted otherwise, all variants are fine-tuned following the RL protocol in Sec.~\ref{sec:exp_scan_rl}.

Table~\ref{tab:spatial_module_ablation} reports few-shot success rates as components are progressively removed from the full model. 
Each row cumulatively removes modules to isolate the effects of spatially conditioned exploration (SCAN), step-level dense rewards (DR), and spatial token fusion.

Removing SCAN leads to a clear performance drop, showing that spatially conditioned exploration is critical for robustness after RL adaptation. 
Further removing dense rewards results in the lowest success among RL-finetuned variants, consistent with prior results on reward density. 
Spatial token fusion alone yields modest gains but cannot ensure robustness without reward shaping and exploration.

The final row shows a non-RL baseline, which lacks task-specific adaptation. 
While it performs better than sparsely rewarded RL fine-tuning, it is not directly comparable to RL-adapted variants.

\begin{table}[ht]
    \centering
    \caption{
        Cumulative ablation of spatial components on few-shot success rate (\%) on LIBERO-PLUS.
        The bottom row is evaluated without RL adaptation, while all other rows report RL-finetuned policies obtained in Sec.~\ref{sec:exp_scan_rl}.
    }
    \label{tab:spatial_module_ablation}
    \begin{tabular}{l c c c c}
        \toprule
        Method & SCAN & DR & Fusion & SR (\%) \\
        \midrule
        SA-VLA & \checkmark & \checkmark & \checkmark & \textbf{83.75} \\
        w/o SCAN & $\times$ & \checkmark & \checkmark & 83.00 \\
        w/o DR & $\times$ & $\times$ & \checkmark & \textcolor{gray}{77.50} \\
        w/o Fusion & $\times$ & $\times$ & $\times$ & 81.00 \\
        \bottomrule
    \end{tabular}
\end{table}

Overall, the ablation confirms that SCAN, dense rewards, and spatial fusion provide complementary benefits. 
The strongest performance is achieved when all components are enabled, while removing any subset during RL adaptation systematically degrades spatial robustness.

\section{Conclusion}
\label{sec:conclusion}

We identify that the degradation of flow-matching Vision-Language-Action policies under spatial distribution shifts stems from a collapse of spatial inductive bias during RL fine-tuning. 
To address this, we propose SA-VLA, which preserves spatial awareness through aligned representation learning, reward design, and exploration. 
Across diverse manipulation tasks, this spatially grounded adaptation improves robustness and transferability, highlighting the importance of preserving inductive structure when extending flow-based VLA policies beyond supervised pretraining.




\normalsize
\bibliography{references}

\clearpage
\appendix
\onecolumn

\section{Additional Methodological Insights}
\label{sec:app_method}

This appendix provides complementary methodological details that clarify several design and implementation choices underlying our approach.
Rather than repeating standard optimization settings described in the main text, we focus on aspects that are either implicit in the algorithmic formulation or tightly coupled to the stability of reinforcement learning fine-tuning under spatial distribution shifts.
These details aim to improve clarity and reproducibility while preserving the conceptual flow of the main paper.

\subsection{Why Implicit Spatial Tokens Instead of Explicit 3D Representations}

A natural alternative to spatial tokens is to explicitly reconstruct 3D geometry using point clouds, voxel grids, or signed distance fields.
However, under reinforcement learning fine-tuning of pretrained policies, explicit 3D representations are poorly matched to the optimization objective.

Explicit reconstruction pipelines typically rely on auxiliary supervision or reconstruction losses.
When combined with policy optimization, this introduces competing gradient signals that are not directly aligned with the reinforcement learning objective,
\begin{equation}
    \nabla_\theta \mathbb{E}\!\left[\sum_t r_t \right],
\end{equation}
and can destabilize policy updates.
Moreover, discretized 3D representations are sensitive to partial observability, sensor noise, and viewpoint sparsity, which are exacerbated during online data collection.
These factors jointly lead to high-variance gradients and brittle adaptation behavior.

In contrast, implicit spatial tokens encode geometric structure directly within a learned feature space derived from multi-view observations.
They preserve continuous spatial information while remaining fully differentiable and directly coupled to the policy backbone.
As a result, spatial reasoning is injected into the policy through feature modulation rather than through an explicit reconstruction objective.
This allows geometric cues to influence action selection while remaining fully aligned with the reinforcement learning loss.

We therefore treat spatial tokens as complementary geometric context rather than explicit scene reconstructions.
This design preserves the simplicity of the policy learning pipeline while enabling stable and geometry-aware adaptation under spatial distribution shifts.

\subsection{Directional Design of Spatial Token Fusion}

Spatial token fusion is implemented using unidirectional cross-attention from visual tokens to spatial tokens.
This asymmetric design reflects the distinct roles of the two representations.
Visual tokens encode task-relevant semantics and fine-grained appearance cues, but are sensitive to occlusion and viewpoint variation.
Spatial tokens encode coarser but more stable geometric structure derived from multi-view aggregation.

Unidirectional attention allows visual tokens to selectively query geometric context that is relevant for action generation.
Formally, spatial tokens serve as keys and values that condition visual features, while remaining invariant to gradients originating from appearance noise.
In contrast, bidirectional attention would allow noisy visual updates to directly modify spatial representations, gradually eroding their geometric consistency during reinforcement learning.

By keeping spatial tokens as a read-only context during fusion, the model preserves a stable geometric anchor throughout fine-tuning.
This asymmetric interaction improves robustness under partial observability and prevents destructive interference between appearance-driven and geometry-driven signals.

\subsection{Design Rationale of Channel-wise Gating}

Channel-wise gating regulates how spatial information modulates visual representations during policy adaptation.
Unlike a simple residual addition or a scalar gate, channel-wise modulation allows the model to selectively amplify or suppress geometric cues along different semantic dimensions.
This is critical because only a subset of visual channels benefit from spatial conditioning at any given state.

The gating function is bounded using a hyperbolic tangent activation, which constrains feature modulation to a fixed range.
This prevents excessive amplification of spatial features during early stages of reinforcement learning, where policy gradients can be highly unstable.
At the same time, bounded modulation still permits strong geometric influence in regions affected by occlusion, clutter, or viewpoint changes.

Empirically, this design stabilizes policy optimization while preserving the expressive capacity of the fused representation.
It allows spatial cues to guide action selection without overwhelming pretrained visual semantics.

\subsection{Why Reach-Place-Leave Decomposition}

The Reach-Place-Leave decomposition is not a task-specific script, but a minimal abstraction of object-centric manipulation geometry.
Across a wide range of manipulation tasks, successful execution requires reducing the distance between the end-effector and the target object, transporting the object toward a goal region, and increasing separation after placement.
These stages correspond to distinct geometric relations that can be measured using relative distances.

Phase assignment is inferred online from interaction signals rather than predefined temporal schedules.
This enables policies to transition between phases based on execution dynamics rather than fixed time steps.
As a result, dense supervision is provided without constraining the temporal structure of the policy.

Importantly, the reward formulation depends only on signed changes in relative geometric distances within each phase.
It does not rely on absolute thresholds or hand-designed trajectories.
This preserves flexibility while embedding minimal manipulation priors into the reinforcement learning objective.

\subsection{Why Spatially-Conditioned Exploration Beyond Dense Rewards}

Dense rewards shape the direction of policy gradients but do not directly control state visitation.
In spatially complex environments, policy optimization can converge to locally consistent but globally suboptimal behaviors when exploration fails to cover critical geometric configurations.

Spatially-conditioned exploration addresses this limitation by explicitly biasing action sampling toward spatially meaningful regions of the state space.
While dense rewards encode task progress, exploration governs which geometric interactions are experienced during training.
These two mechanisms therefore operate at different levels of the learning process.

By combining phase-consistent dense rewards with spatially-conditioned exploration, SA-VLA improves both gradient quality and state coverage.
This combination is essential for stable fine-tuning under spatial distribution shifts.

\section{Detailed Algorithmic and Implementation Details}
\label{sec:app_impl}

This section provides additional algorithmic and implementation details that complement, rather than replicate, the main method description.
While the core model design and learning objectives are presented in Secs.~\ref{sec:fuser}-\ref{sec:scan}, several practical choices are difficult to fully articulate in the main text without interrupting the methodological narrative.

We therefore focus on implementation-level considerations that are critical for training stability, numerical robustness, and reproducibility, including token handling strategies, phase inference logic, reward normalization, and exploration scheduling.
These details reflect constraints encountered during online reinforcement learning and are essential for faithfully reproducing the reported results.

\subsection{Choice and Analysis of Exploration Noise}
\label{sec:app_noise_analysis}

The parameterization of exploration noise, in particular whether stochasticity is modeled as an intrinsic component of the policy or introduced externally, plays a critical role in both the theoretical validity of PPO optimization and the interpretation of experimental results. Accordingly, we employ different exploration noise mechanisms across experiments to isolate specific factors while maintaining conceptual clarity.

\begin{itemize}
    \item \textbf{SDE-based isotropic noise (SDE):} used in Sec.~\ref{sec:exp_reward_density_spatial} to isolate the effect of reward density. This geometry-agnostic noise serves as a controlled baseline, enabling a fair comparison between sparse and dense rewards without introducing policy-dependent exploration.
    
    \item \textbf{Flow\_noise:} used in Sec.~\ref{sec:exp_scan_rl} to evaluate spatially-conditioned exploration. This learned noise head generates geometry-consistent stochasticity from spatial tokens and is fully parameterized as part of the flow-matching policy, ensuring compatibility with PPO optimization.
    
    \item \textbf{SCAN (Spatially-Conditioned Annealed Noise):} combines a learned, geometry-consistent noise head with a temporarily injected isotropic SDE floor. The annealing schedule ensures that exploration noise converges to a policy-dependent distribution, reconciling early-stage exploration with theoretical consistency.
\end{itemize}

\paragraph{Mathematical analysis under PPO}  
We formalize the distinction by considering a Gaussian policy
\[
\pi_\theta(x_{t-\Delta} \mid x_t, s) = \mathcal{N}\big(\mu_\theta(x_t, s), \sigma_\theta(x_t, s)^2 \big),
\]
where \(x_{t-\Delta} = \mu_\theta(x_t, s) + \sigma_\theta(x_t, s) \cdot \epsilon\) and \(\epsilon \sim \mathcal{N}(0,1)\).

\begin{enumerate}
    \item \textbf{SDE-based isotropic noise} injects stochasticity via
    \[
    x_{t-\Delta} = \mu_\theta(x_t) + \sigma(t)\sqrt{\Delta}\,\epsilon, \quad
    dx = v_\theta(x,t)\, dt + \sigma(t)\, dW_t,
    \]
    where \(dW_t\) is a Brownian increment and \(\sigma(t)\) is independent of policy parameters \(\theta\).
    As a result:
    \begin{itemize}
        \item The injected noise is \emph{external} to the policy \(\pi_\theta\), and therefore not reflected in the likelihood ratio optimized by PPO.
        \item This leads to an objective mismatch, as the effective action distribution used for exploration differs from the distribution assumed in PPO gradient estimation.
        \item Consequently, variance cannot adapt to the reward signal; only the mean is optimized, limiting the exploration-exploitation trade-off and potentially amplifying instability under sparse or long-horizon rewards.
    \end{itemize}
    
    \item \textbf{Flow\_noise} instead defines
    \[
    x_{t-\Delta} = \mu_\theta(x_t, s) + \sigma_\theta(x_t, s) \cdot \epsilon,
    \]
    where \(\sigma_\theta\) is learned from spatial tokens and explicitly parameterized as part of the policy.
    This ensures that both the mean and variance contribute to the likelihood ratio \(r_t(\theta)\), making surrogate objectives such as PPO's clipped loss
    \[
    L^{\text{CLIP}}(\theta) =
    \mathbb{E}_t \Big[
        \min \big(
            r_t(\theta) \hat{A}_t,\,
            \text{clip}(r_t(\theta), 1-\epsilon, 1+\epsilon)\hat{A}_t
        \big)
    \Big]
    \]
    theoretically consistent.
    
    \item \textbf{SCAN} temporarily injects SDE-style noise to encourage early exploration, but anneals it toward the learned \(\sigma_\theta(x_t, s)\).
    After annealing, all stochasticity is policy-dependent, guaranteeing that the final optimization objective remains mathematically aligned with PPO.
\end{enumerate}

\begin{figure}[h]
    \centering
    \includegraphics[width=0.45\textwidth]{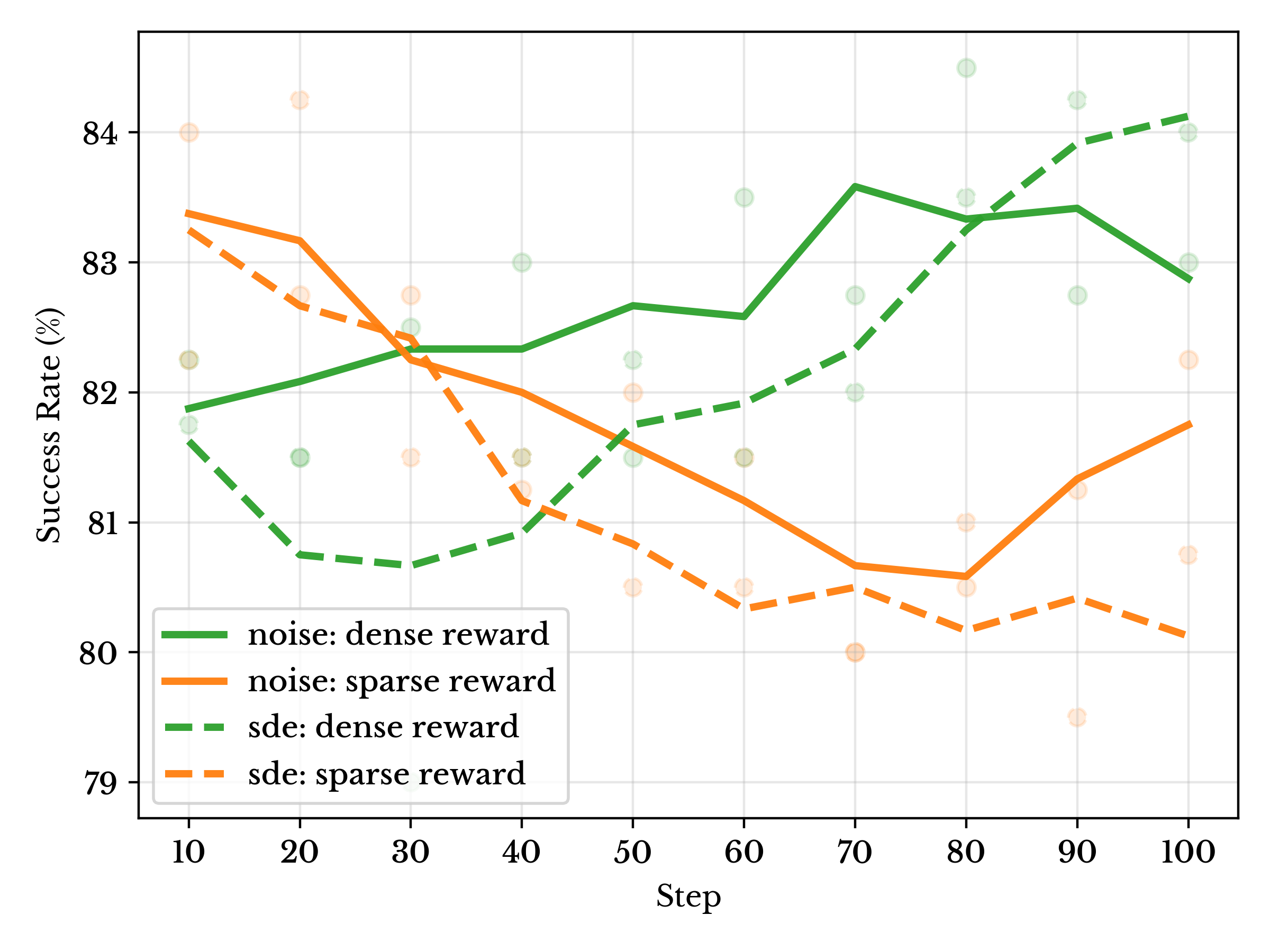}
    \caption{
        Few-shot evaluation on LIBERO-PLUS comparing SDE-based and learned exploration noise.
        Across both sparse and dense reward settings, learned noise consistently outperforms SDE-based noise with lower variance and more stable performance, highlighting the robustness of policy-dependent exploration.
    }
    \label{fig:noise_comparison}
\end{figure}

\paragraph{Empirical comparison}  
Figure~\ref{fig:noise_comparison} reports few-shot evaluation on LIBERO-PLUS under dense and sparse rewards.
Notably, the advantage of learned noise is consistent across reward regimes, indicating that policy-dependent stochasticity improves exploration robustness rather than merely amplifying reward density.

\paragraph{Discussion}  
\begin{itemize}
    \item Flow\_noise and SCAN exploit spatial priors by learning exploration variance from fused visual-spatial tokens, enabling the policy to modulate stochasticity based on geometry-aware state representations.
    \item SDE-based noise is suitable for isolating reward-density effects, but is theoretically inconsistent with PPO due to its policy-independent variance.
    \item SCAN reconciles effective early-stage exploration with mathematical rigor by annealing external noise toward a policy-dependent distribution.
    \item By conditioning exploration noise on fused spatial tokens, learned noise preserves spatial inductive bias during RL adaptation, mitigating shortcut learning that arises from geometry-agnostic stochasticity.
    \item Empirically, learned noise yields more stable and consistent behavior, supporting the importance of policy-dependent, geometry-aware stochasticity.
\end{itemize}

Overall, this analysis clarifies that comparisons across reward density and exploration strategies are only meaningful when the exploration noise is parameterized consistently with the underlying policy, thereby justifying our experimental design choices.

These observations justify our experimental design: SDE for reward density analysis, and flow\_noise / SCAN for evaluating spatially-conditioned exploration beyond dense rewards.

\subsection{Implementation Details of Spatial Token Fusion}
\label{sec:app_impl_fuser}

This subsection clarifies several implementation-level design choices in the spatial token fusion module that are not explicitly discussed in the main text.

\paragraph{Separation of grid and global spatial tokens.}
The spatial encoder produces two types of tokens: dense grid tokens encoding spatial layout, and a small set of global tokens summarizing scene-level geometry.
Only grid tokens participate in cross-attention with visual tokens.
Global tokens are concatenated after fusion and bypass the attention mechanism.

This separation is intentional.
Grid tokens provide localized geometric cues suitable for selective querying by visual semantics, while global tokens encode coarse context that does not benefit from fine-grained alignment.
Allowing global tokens to enter cross-attention was empirically observed to introduce noisy gradients and degrade training stability.

\paragraph{Read-only treatment of spatial tokens.}
Spatial tokens are frozen during reinforcement learning and serve as keys and values in cross-attention.
Gradients from policy updates do not modify spatial features, preventing appearance-driven noise from corrupting geometric structure.
This design preserves the role of spatial tokens as a stable geometric reference throughout fine-tuning.

\subsection{Stability-Oriented Phase Inference}
\label{sec:app_impl_phase}

Phase inference is implemented using stability-based criteria rather than distance thresholds.
This choice avoids premature phase transitions caused by noisy observations or transient contacts.

Specifically, phase transitions depend on short-horizon consistency of relative poses and gripper state, rather than instantaneous distance values.
This prevents oscillations between Reach and Place when the end-effector briefly contacts the object, and avoids false Leave transitions due to transient object motion.

When a phase switch occurs, the reference distance for the newly active phase is reinitialized.
This prevents spurious reward spikes at transition boundaries and ensures that dense rewards reflect local geometric progress within each phase.

\subsection{Reward Scaling and Numerical Stability}
\label{sec:app_impl_reward}

Dense rewards are computed as temporal differences of normalized distances.
Normalization by the initial valid distance removes dependence on absolute scene scale and yields comparable reward magnitudes across tasks.

We additionally clip normalized distances to the unit interval and optionally clip reward values.
These operations prevent large gradients caused by sudden distance changes, especially during early exploration or failed grasps.

Empirically, we observe that performance is robust to moderate variations of the scaling coefficient.
This indicates that dense rewards primarily act as a directional geometric bias rather than a finely tuned optimization objective.

\begin{figure}[t]
\centering
\includegraphics[width=0.16\linewidth]{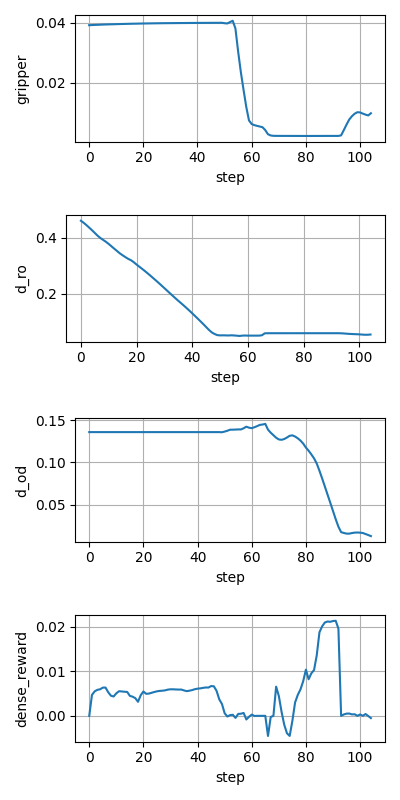}
\includegraphics[width=0.16\linewidth]{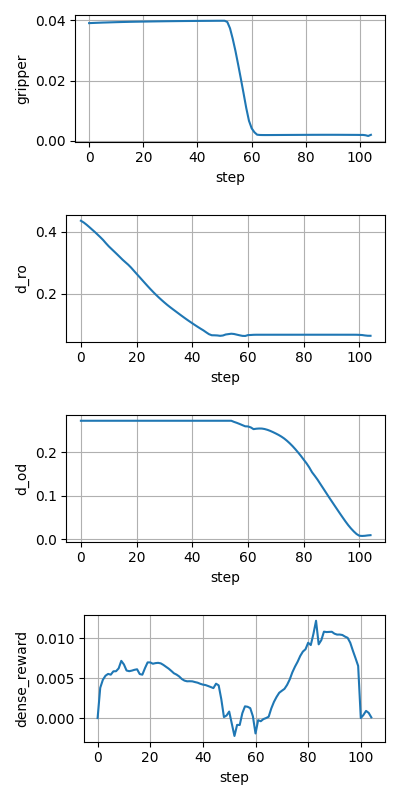}
\includegraphics[width=0.16\linewidth]{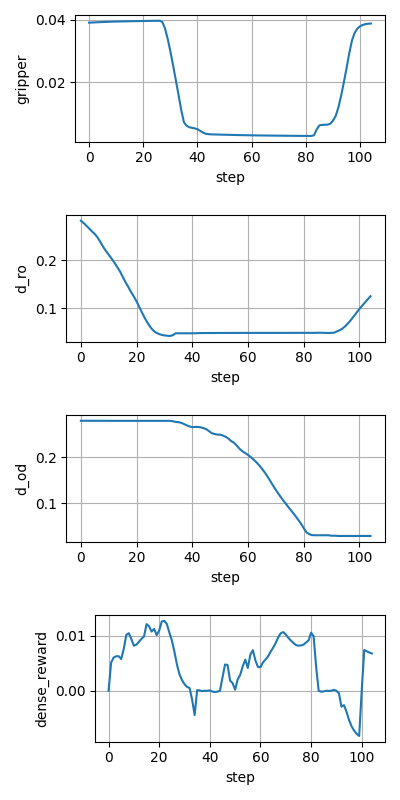}
\includegraphics[width=0.16\linewidth]{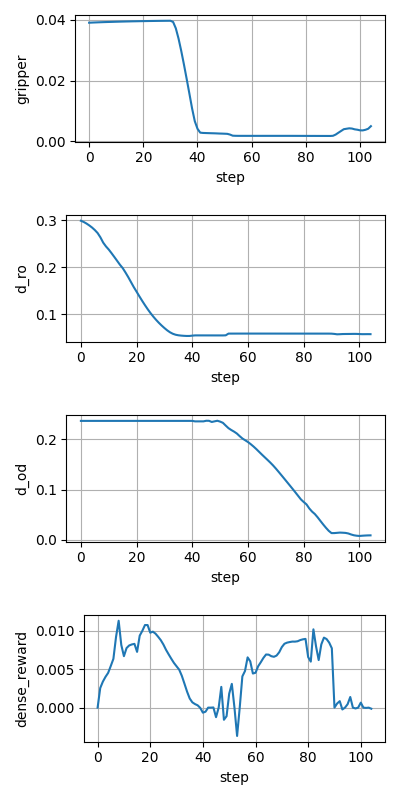}
\includegraphics[width=0.16\linewidth]{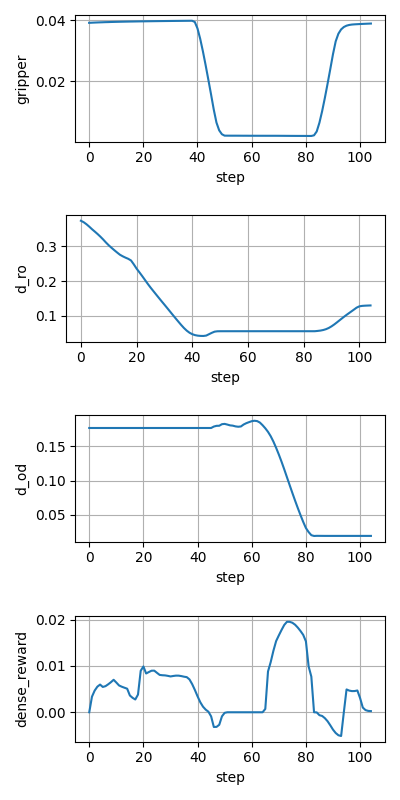}
\includegraphics[width=0.16\linewidth]{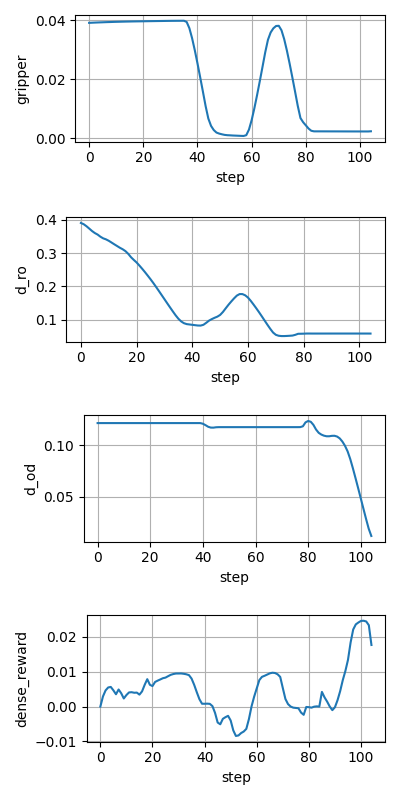}
\caption{
    \textbf{Phase-wise dense reward visualization.}
    Shown are changes in $d_\text{ro}$, $d_\text{od}$, gripper opening angle, and the corresponding dense reward throughout task execution.
}
\label{fig:phase_rewards}
\end{figure}

\section{Additional Experimental Results}
\label{sec:app_additional}

To further illustrate the efficacy of SA-VLA, we provide additional qualitative and interpretative analyses that complement the quantitative results presented in the main text. 
These results highlight the robustness of the policy under spatial perturbations and the interpretability of the step-level dense rewards.

\subsection{Phase-wise Reward Visualization}

To illustrate the behavior of the dense reward, Fig.~\ref{fig:phase_rewards} closely tracks geometric progress: it increases when the end-effector approaches or moves the object toward the target, and decreases when actions deviate from the intended motion. 
The reward dynamics qualitatively reflect the underlying manipulation phases, providing interpretable learning signals aligned with task progression.

\subsection{Qualitative Results Under Spatial Perturbations}
\label{sec:app_eval}

\begin{figure}[ht]
\centering
\includegraphics[width=\linewidth]{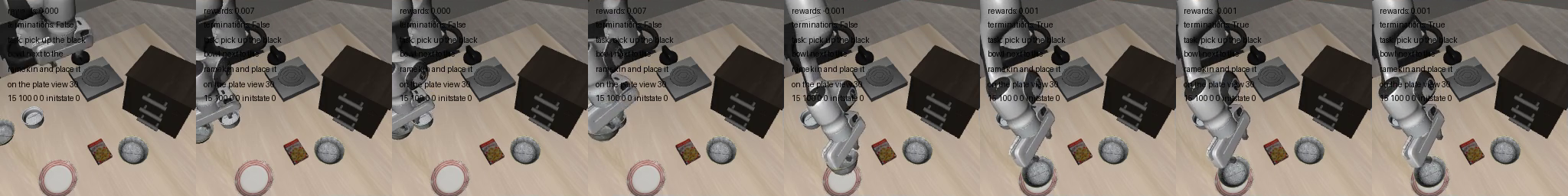}\\[1ex]
\includegraphics[width=\linewidth]{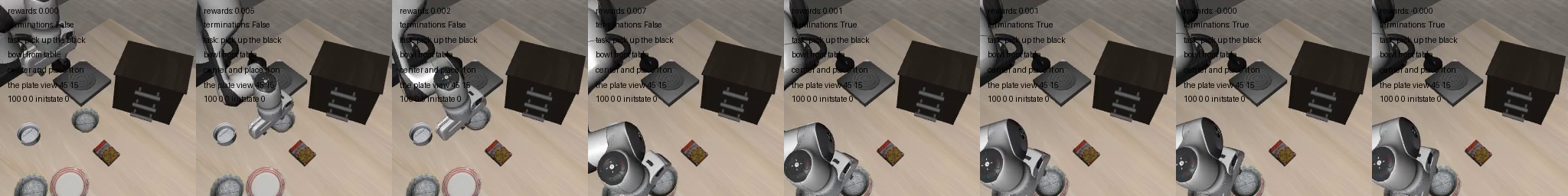}\\[1ex]
\includegraphics[width=\linewidth]{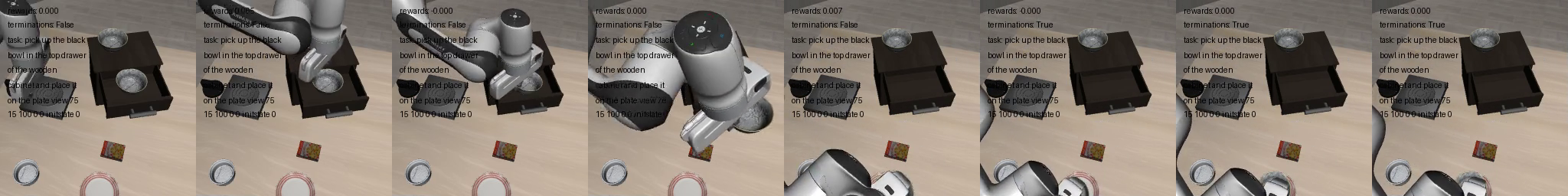}\\[1ex]
\includegraphics[width=\linewidth]{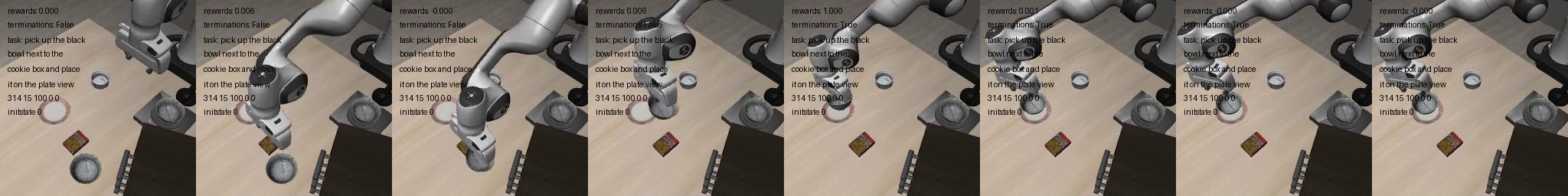}\\[1ex]
\includegraphics[width=\linewidth]{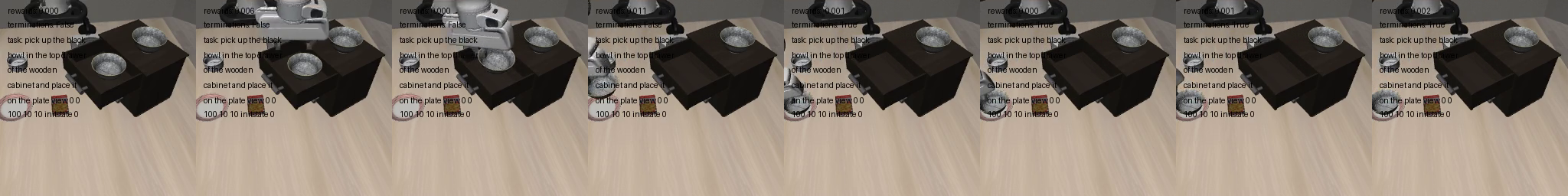}\\[1ex]
\includegraphics[width=\linewidth]{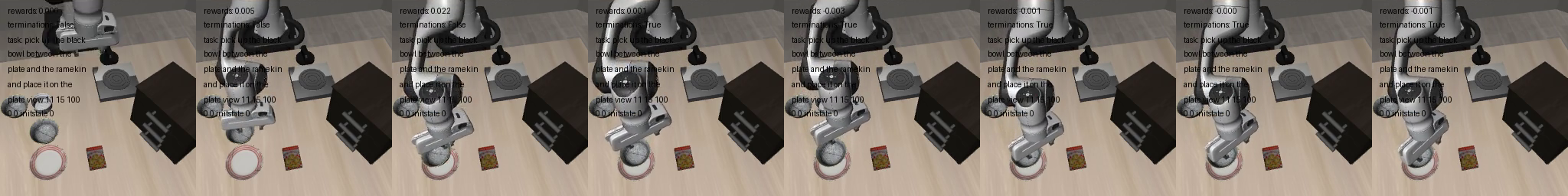}\\[1ex]
\includegraphics[width=\linewidth]{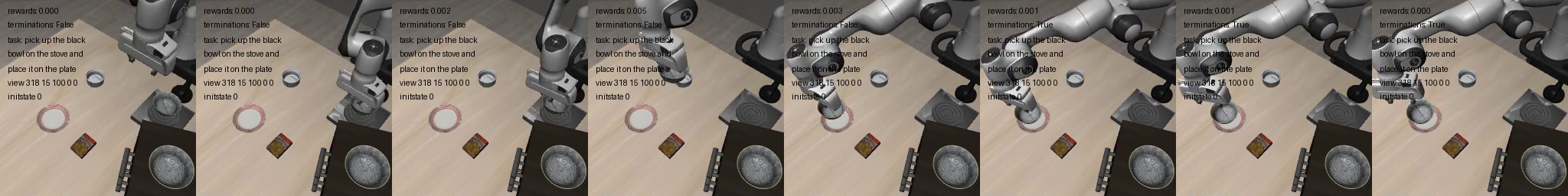}
\caption{
\textbf{Representative task executions under spatial perturbations.} 
Keyframes are uniformly sampled (8 per rollout) from successful episodes. Despite variations in observation geometry, the policy consistently achieves task completion.
}
\label{fig:long}
\end{figure}

We visualize representative rollouts under diverse spatial perturbations in Fig.~\ref{fig:long}, sampling 8 frames uniformly along each episode horizon. 
All perturbations are applied at evaluation time. 
Across settings, the policy reliably progresses toward the target and completes the task, supporting the robustness trends observed in quantitative evaluations.

\section{Limitations and Discussion}
\label{sec:app_limit}

While SA-VLA generalizes across a range of object-centric manipulation tasks, several limitations remain. 
The Reach-Place-Leave abstraction captures common interactions but may require adaptation for tasks involving deformable objects or extended contact. 
The quality of spatial tokens is dependent on the pretrained spatial encoder, which can constrain performance in visually ambiguous or cluttered scenes. 
Future work could explore alternative interaction modalities, richer geometric representations, and more flexible phase abstractions to extend applicability and robustness.

\end{document}